\documentclass[10pt,twocolumn,letterpaper]{article}
\usepackage[pagenumbers]{cvpr}
\usepackage{fontawesome5}
\usepackage{currfile} %< for \currfilebase macro

\usepackage{caption} %< for "\captionof" commands (teaser)
\usepackage{xspace}
\newcommand{\pvp}{CriticJEPA\xspace}

\DeclareMathOperator*{\argmin}{arg\,min}

\usepackage{amsmath,amssymb}
\usepackage{booktabs}
\usepackage{array}
\usepackage{multirow}
\usepackage{graphicx}
\usepackage{tikz}
\usetikzlibrary{arrows.meta,calc,positioning}
\definecolor{pvpnavy}{HTML}{1F2937}
\definecolor{pvpblue}{HTML}{3B6FB6}
\definecolor{pvpteal}{HTML}{2E7D6B}
\definecolor{pvpamber}{HTML}{C17F24}
\definecolor{pvpred}{HTML}{C0504D}
\definecolor{pvpgray}{HTML}{6B7280}
\definecolor{pvppanelA}{HTML}{F5F6F3}
\definecolor{pvppanelB}{HTML}{F1F6F4}
\definecolor{pvpfrozen}{HTML}{EEF0F8}
\definecolor{pvpfrozenD}{HTML}{6E79A8}
\usepackage{microtype}
\newcommand{\ds}[1]{{\scriptsize$\pm$#1}}

\usepackage{flushend}

\definecolor{cvprblue}{rgb}{0.21,0.49,0.74}
\definecolor{projectpink}{HTML}{A71F67}
\usepackage[breaklinks,colorlinks,allcolors=cvprblue]{hyperref}

\makeatletter
\patchcmd{\@maketitle}{\vskip .375in}{\vskip 16pt}{}{}
\patchcmd{\@maketitle}{\vspace*{24pt}}{\vspace*{18pt}}{}{}
\makeatother

\title{CriticJEPA: World Models Are Better Critics Than Optimizers}
\author{%
  Xvyuan Liu\textsuperscript{1,*}, Jianjie Fang\textsuperscript{1,*},
  Honglin Zhang\textsuperscript{1,*}, Zhaolu Wang\textsuperscript{2},
  Xin Zhang\textsuperscript{3},\\
  Wei Wu\textsuperscript{3}, Fengli Xu\textsuperscript{1},
  Chen Gao\textsuperscript{1,\ensuremath{\dagger}},
  Yong Li\textsuperscript{1,\ensuremath{\dagger}}\\[0.45em]
  {\normalsize \textsuperscript{1}Tsinghua University\quad
   \textsuperscript{2}Zhongguancun Academy\quad
   \textsuperscript{3}Manifold AI}\\[0.25em]
  {\small \textsuperscript{*}Equal contribution\qquad
   \textsuperscript{\ensuremath{\dagger}}Corresponding authors}\\[0.55em]
  {\normalsize\itshape
   \href{https://daybraeklaxry.github.io/criticjepa/}{\textcolor{projectpink}{\faGlobe\enspace Project Page}}
   \qquad
   \href{https://github.com/daybraeklaxry/CriticJEPA-code}{\textcolor{black}{\faGithub\enspace Code}}}
}

\begin{document}
\raggedbottom
\twocolumn[{
\renewcommand\twocolumn[1][]{#1}
\maketitle
\begingroup
\centering
\includegraphics[width=\linewidth]{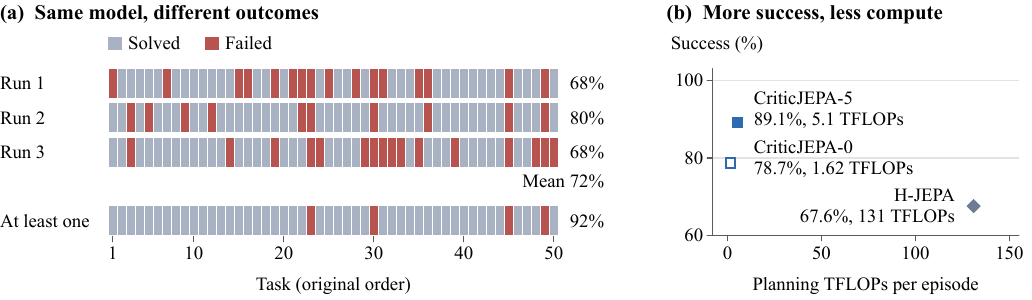}
\captionsetup{hypcap=false}
\captionof{figure}{\textbf{\pvp improves success with less planning compute.} \textbf{(a)}~Three H-JEPA restarts on the same 50 AntMaze tasks in original order. Mean single-run success is 72\%; the union row covers 92\%. \textbf{(b)}~Success and compute across three independently trained models on Standard AntMaze. Suffixes denote refinement steps per level; candidate counts follow \cref{tab:antmaze}.}
\label{fig:teaser}
\par
\endgroup

\bigskip
}]
% Discard the template's abstract spacing at a column break.
\patchcmd{\endabstract}{\vspace*}{\vspace}{}{}
\begin{abstract}
Visual world models enable goal-directed planning by predicting action outcomes. Gradient-based planners optimize actions to lower predicted goal cost, but continued optimization can exploit prediction errors after execution stops improving. Comparing plans requires informative relative scores; improving them also requires reliable update directions. Exact-state execution reveals that a frozen hierarchical world model ranks goal-conditioned proposals even when its rankings among outputs of prolonged gradient search are uninformative. Effective planning pairs goal-directed candidates with predictions that distinguish their outcomes. We introduce \emph{\mbox{\pvp}}, which learns goal-conditioned flow proposals from reward-free trajectories and uses frozen rollouts to select among them at every hierarchy level. Forward selection forms a complete planner; short anchored updates refine candidates before selection. Across visual navigation and manipulation, \pvp increases success at a fraction of the planning compute, transferring shared settings across independently trained models, hierarchy depths, and unseen tasks while keeping every world-model weight fixed.
\par
\end{abstract}

\section{Introduction}
\label{sec:intro}

Visual world models predict how the environment changes under a sequence of actions, enabling agents to plan toward goal images in a learned latent space~\cite{lecun2022path,zhou2025dinowm,sobal2025pldm,assran2025vjepa2,maes2026lewm}. Gradient-based planners use these predictions both to evaluate plans and to construct them, repeatedly updating candidate actions to reduce predicted goal cost. H-JEPA~\cite{zhang2026hjepa} extends this process across temporal scales, from high-level subgoals to low-level controls.

With the world model fixed, H-JEPA restarts solve different sets of tasks, as shown in \cref{fig:teaser}a. The same planning stack reaches more goals across runs than a single search reliably recovers. Tracing the action updates reveals a mismatch between predicted improvement and execution: early updates improve both, but further updates lower predicted endpoint error after execution stalls, as shown in \cref{fig:probes}a. Repeated optimization can exploit prediction errors, favoring actions that look increasingly successful to the model without moving closer to the goal in execution.

Evaluating plans and optimizing them require different properties from the predictor. Choosing among given plans requires scores that preserve their relative quality. Improving a plan through gradients also requires reliable directions as the actions change. Errors that leave the ordering of a candidate set intact can still redirect optimization toward artificially favorable predictions. A predictor can distinguish promising plans even when further optimization stops improving their execution.

This distinction appears when we execute alternative plans from identical simulator states, holding the goal and horizon fixed. The same frozen predictor gives informative rankings of learned goal-conditioned proposals and near-chance rankings of H-JEPA search outputs, as shown in \cref{tab:probe}. Goal-conditioned proposals yield better selected outcomes than retrieved training actions: retrieval matches the starting latent state, whereas the proposer also uses the current goal to generate alternatives. A critic's value depends on both how well it compares plans and which plans \mbox{it is given}.

These observations lead to \emph{\pvp}, which separates the generation of plans from their evaluation. Goal-conditioned flow heads~\cite{lipman2023flow,liu2023rectified} learn action sequences from reward-free trajectories paired with the states they actually reach~\cite{andrychowicz2017her,ghosh2021gcsl}. Frozen world-model rollouts then select among the generated alternatives. Short gradient updates capture the early gains in execution, with an anchor penalizing departure from each proposal.

The design operates throughout the temporal hierarchy, as illustrated in \cref{fig:method}: selected upper-level plans supply subgoals, and bottom-level plans supply controls. Forward selection constructs complete plans; the refinement variant updates every candidate before selection. In contrast to prior-guided iterative search in flat JEPA models~\cite{wang2026prism,cheng2026sage}, generation and frozen selection form the planner at every level, including the macro-actions that determine \mbox{subsequent subgoals}.

Across visual navigation and manipulation, \pvp reaches more goals with substantially less planning compute, as shown for AntMaze in \cref{fig:teaser}b. Our contributions are:
\begin{itemize}
    \item We demonstrate through exact-state AntMaze execution that a frozen predictor distinguishes goal-conditioned proposals when prolonged gradient search produces uninformative rankings. The candidate source determines how ranking translates into executed outcomes.
    \item We introduce \pvp, a complete hierarchical planner that pairs goal-conditioned flow proposals with frozen rollout selection and short anchored updates for \mbox{local refinement}.
    \item We demonstrate success and compute gains across visual navigation and manipulation, transferring shared planning settings across independently trained models, hierarchy depths, and unseen AntMaze tasks.
\end{itemize}

\section{Related Work}
\label{sec:related}

\paragraph{Planning with self-supervised world models.}
Learning latent dynamics enables planning from high-dimensional visual observations~\cite{hafner2019planet,chua2018pets,hansen2024tdmpc2}. Joint-Embedding Predictive Architectures (JEPA)~\cite{lecun2022path,assran2023ijepa,bardes2024vjepa,balestriero2025lejepa} learn predictive representations without reconstructing pixels. Visual world models including PLDM~\cite{sobal2025pldm}, DINO-WM~\cite{zhou2025dinowm}, V-JEPA~2~\cite{assran2025vjepa2}, NWM~\cite{bar2025nwm}, and LeWorldModel~\cite{maes2026lewm} support goal-image planning from reward-free trajectories; Terver \etal~\cite{terver2026what} study the design choices behind such planners. HWM~\cite{zhang2026hwm} and H-JEPA~\cite{zhang2026hjepa} extend latent prediction across temporal scales. With pretrained H-JEPA hierarchies, we show that changing how plans are generated and judged improves control without changing the learned dynamics.

\paragraph{Model exploitation in trajectory optimization.}
Optimizing actions against learned dynamics can favor out-of-distribution inputs on which predictions are erroneously optimistic~\cite{chua2018pets,janner2019mbpo}, paralleling adversarial examples for classifiers~\cite{szegedy2014intriguing}. In a two-level LeWorldModel, Mind the Gap~\cite{caselli2026mindthegap} observes that CEM over macro-actions exploits the high-level predictor and regularizes search toward the data. Our exact-state execution shows that a frozen predictor distinguishes goal-conditioned proposals when rankings among its gradient-search outputs are uninformative. The candidate source determines how these comparisons translate into the quality of the selected action.

\paragraph{Generative priors and amortized proposals.}
Policy priors accelerate model-based search~\cite{wang2020poplin,hansen2024tdmpc2}, while diffusion models generate trajectories and actions~\cite{janner2022diffuser,ajay2023decisiondiffuser,chi2023diffusionpolicy}. In flat JEPA models, PRISM~\cite{wang2026prism} adds a Gaussian prior to MPPI and SAGE~\cite{cheng2026sage} initializes CEM with learned subgoals and actions. LePlanner~\cite{bansal2026leplanner} amortizes planning with a recurrent controller, and INTACT~\cite{sun2026intact} maps latent goal displacements to actions. In \pvp, proposal generation and frozen rollout selection extend throughout the temporal hierarchy, including the macro-actions that determine subgoals. Forward evaluation alone supports planning at every level, and anchored refinement spends further computation on short corrections to the proposals.

\section{Diagnosing the Search Bottleneck}
\label{sec:probing}

Planning with a learned predictor couples the model's predictions to the actions produced by search. We separate these effects through exact-state execution with one frozen three-level H-JEPA hierarchy~\cite{zhang2026hjepa} on Visual AntMaze. Upper levels choose subgoals, and the bottom level supplies controls. The H-JEPA planner initializes candidates from Gaussian distributions and minimizes predicted goal-distance cost with up to 90 AdamW steps~\cite{loshchilov2019adamw}.

\subsection{Predicted Improvement, Stalled Execution}
\label{sec:exploit}

At each saved state, we fix the current level-$m$ subgoal $g^{(m)}$. For each candidate, we predict and execute the same replanning window: 10 primitive steps, or 5 at the shortest horizon. Let $\tilde z_{\mathrm{end},i}^{(m)}$ be the predicted endpoint projected through frozen encoders into the target space, and $E^{(m)}(o_{\mathrm{end}}^{(i)})$ the reached image encoded in that space. \mbox{The window-level diagnostic compares}
\begin{equation}
\begin{aligned}
C_{\mathrm{pred},i}^{\mathrm{win}}&=\|\tilde z_{\mathrm{end},i}^{(m)}-g^{(m)}\|_2^2,\\
C_{\mathrm{true},i}^{\mathrm{win}}&=\|E^{(m)}(o_{\mathrm{end}}^{(i)})-g^{(m)}\|_2^2.
\end{aligned}
\label{eq:window-error}
\end{equation}
The simulator is reset to the same full state before each execution. These endpoint errors measure progress over the replanning window; the planner optimizes the full-horizon cost in \cref{sec:method}.

\Cref{fig:probes}a tracks the gradient planner across iterations $k \in \{0, 10, 30, 90\}$ on 720 replanning states. True executed error falls from 25.3 to 23.0 during the first 10 steps, then stalls at 22.8 after 30 steps and 23.4 after 90. Predicted error keeps dropping, while optimism nearly doubles from 8.6 to 16.7. The extra optimization chiefly improves the model score rather than the executed outcome.

\paragraph{Comparison and optimization impose different requirements.}
For matched predicted and realized costs, write $C_{\mathrm{pred}}(a)=C_{\mathrm{true}}(a)-\epsilon(a)$, where $\epsilon$ denotes optimism. A local differentiable view gives
\begin{equation}
-\nabla_a C_{\mathrm{pred}}(a)
=-\nabla_a C_{\mathrm{true}}(a)+\nabla_a\epsilon(a).
\label{eq:optimism}
\end{equation}
The update direction contains a term that improves the true objective and a term that increases model optimism. This gives search a route to lower predicted cost without better execution, matching the divergence in the iteration probe.

Comparison makes a different demand. On a fixed candidate pool, a shared prediction-error offset leaves every ranking unchanged. What matters is the error difference between the supplied plans. Gradient search also responds to how the error changes as actions move; even small value errors can vary sharply. Ranking given plans and supplying update directions are distinct predictive abilities.

\subsection{Candidate Generation Shapes Selection}
\label{sec:asym}

This difference between comparison and optimization makes candidate generation central to model-based selection. We execute plans from different sources at identical saved simulator states, with the same subgoal and horizon, and compare the predicted and realized endpoint errors.

\begin{table}[t]
\centering
\small
\caption{\textbf{Goal-conditioned proposals yield the best selected actions.} At each of 300 AntMaze states, eight candidates per source are executed from the same simulator state. \emph{Rank $\rho$}: within-state Spearman correlation of predicted and realized endpoint errors. \emph{Top-1}: how often the model selects the best of the eight; chance is 12.5\%. \emph{Selected error}: mean realized error of the selected action. \emph{Mean optimism}: realized minus predicted endpoint error, averaged over all candidates.}
\label{tab:probe}
\vspace{-0.4em}
\renewcommand{\arraystretch}{1.12}
\setlength{\tabcolsep}{3.4pt}
\begin{tabular}{@{}lcccc@{}}
\toprule
\multirow{2}{*}{Candidates} & \multicolumn{2}{c}{Ranking} & \multirow{2}{*}{\shortstack{Selected\\error$\downarrow$}} & \multirow{2}{*}{\shortstack{Mean\\optimism$\downarrow$}} \\
\cmidrule(lr){2-3}
 & Rank $\rho\uparrow$ & Top-1 (\%)$\uparrow$ & & \\
\midrule
Optimized & $-0.03$ & 12.0 & 24.3 & 17.4 \\
Dataset & \textbf{0.41} & \textbf{27.0} & 30.1 & 1.9 \\
Flow proposals & 0.28 & 22.0 & \textbf{23.0} & \textbf{1.2} \\
\bottomrule
\end{tabular}
\vspace{-0.5em}
\end{table}

We sample 15 replanning states from each of 20 validation episodes, saving the full MuJoCo physics state and fixing the current level-$m$ subgoal $g^{(m)}$ at all 300 states. We evaluate eight level-1 candidates from each of three sources. \textbf{Gradient search} supplies the winners of eight independent H-JEPA AdamW solves, each with 100 initializations and up to 90 steps. \textbf{Training dataset} supplies eight recorded action sequences from the 500 nearest-neighbor level-1 latent states in the training set. \textbf{Flow proposals} are eight direct samples from the goal-conditioned proposer in \cref{sec:method}.

The frozen predictor distinguishes candidates from recorded trajectories, as shown in \cref{tab:probe}. Its rankings correlate with executed outcomes at $\rho{=}0.41$, and top-1 selection reaches 27.0\%, over twice the chance rate. On H-JEPA search outputs, ranking correlation is $\rho{=}{-}0.03$ and top-1 selection is at chance level at 12.0\%. Mean optimism, averaged over all candidates, rises from 1.9 for recorded actions to 17.4 for search outputs. The predictor's ability to compare actions thus survives on a different candidate distribution.

Goal-conditioned proposals deliver the best executed choices and the lowest mean optimism: 23.0 and 1.2, respectively. Flow proposals combine informative rankings at $\rho=0.28$ with lower selected error than retrieval, which yields $\rho=0.41$ and error 30.1. Retrieval matches only the starting latent state, whereas the flow also conditions on the current goal. Ranking agreement summarizes the whole pool; control depends on what the selected action achieves. Candidate generation must therefore support both informative comparisons and progress toward the target. 

The selected outcome separates into the best executed cost available in the pool and the extra cost of choosing another candidate. For a fixed pool, the selected plan has no higher predicted cost than the true best plan, so the extra selection cost is bounded by the range of prediction errors across the pool. A shared prediction bias cancels in comparisons, leaving the error differences that affect which action is selected. This gives proposal learning and model evaluation complementary roles: supplying goal-directed alternatives and preserving the comparisons that identify which to execute.

\begin{figure}[t]
\centering
\includegraphics[width=\linewidth]{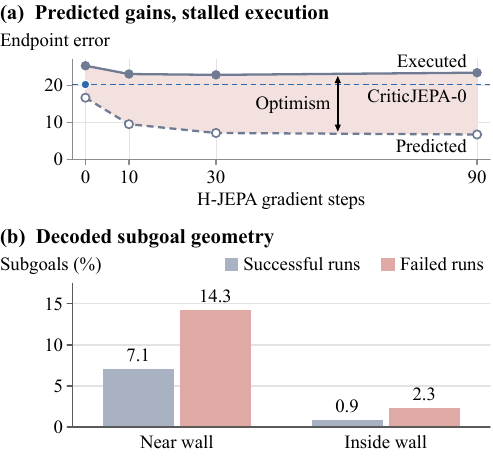}
\vspace{-0.5em}
\caption{\textbf{Optimizing predictions can diverge from execution.} \textbf{(a)}~Predicted and executed endpoint errors at 720 AntMaze states with one frozen model. Shading shows optimism: executed minus predicted error. The blue reference selects among 100 zero-gradient proposals. \textbf{(b)}~Decoded L3 subgoals on the 24 tasks whose outcomes change across restarts. Near-wall subgoals lie within $0.8$\,m of a wall.}
\label{fig:probes}
\vspace{-0.6em}
\end{figure}

Forward selection among 100 zero-gradient \pvp proposals achieves a true error of 20.2 on the 720 states of \cref{fig:probes}a, below every tested step count of H-JEPA \mbox{gradient search}.

\subsection{Search Variability Across the Hierarchy}
\label{sec:gap}

The consequences of search extend through the hierarchy: upper-level choices set targets for lower-level control, so changing either search can change task completion. With the frozen model and initial states held fixed, three H-JEPA restarts on the 50 official tasks produce the outcomes in \cref{fig:teaser}a. Each run has a 350-step budget and independently resampled initial actions. Individual runs solve 68\%, 80\%, and 68\% of tasks, averaging 72.0\%, while their union reaches 46 of 50 tasks, or 92.0\%. Twenty-two tasks are solved by all three runs, and 48\% change outcome across restarts. The gap between individual success and combined coverage identifies a consistency problem: the unchanged planning stack reaches many more goals than a typical run finds. The challenge is to recover this successful behavior reliably from the same frozen model.

Increasing the level-1 pool fivefold to $N_1{=}500$ gives 70.0\% mean success across three further runs. Combining all six runs expands coverage to 48 of 50 tasks, or 96.0\%. Successful behavior is recovered across many tasks by the unchanged planning stack, but a larger initial pool alone leaves single-run success near its original level.

\paragraph{Both high-level and low-level search flip task outcomes.}
Varying search randomness separately at each level reveals instability at both timescales. With lower-level random draws fixed, changing only top-level initialization flips outcomes on 44\% of the 50 tasks. With upper-level draws fixed, changing only bottom-level initialization flips 38\%. The outcome changes at both timescales connect candidate generation to the complete control loop: the upper search chooses where to go next, and the lower search determines how to get there.

At the top level, unsuccessful search has a geometric signature. We train a linear probe on real observations to decode level-3 latents into maze coordinates, obtaining $R^2\approx0.998$. \Cref{fig:probes}b compares successful and failed runs on the 24 tasks whose outcomes change across restarts. In failed runs, 14.3\% of subgoals decode within $0.8$\,m of a wall and 2.34\% inside walls. The corresponding rates for successful runs are 7.1\% and 0.89\%. Failed runs thus more often produce subgoals decoded near or inside walls. Across all 50 tasks, \pvp-0 reduces the fraction decoded inside walls from 2.31\% to 0.31\%. Coordinates are used for this geometric evaluation; training and planning operate in the frozen latent and action spaces.

The hierarchy therefore needs goal-directed alternatives for both subgoal selection and primitive control. Recorded trajectories provide state-to-goal connections at each timescale, from which a generator can learn the plans that the frozen predictor will compare.

\begin{figure*}[t]
\centering
\includegraphics[width=\textwidth]{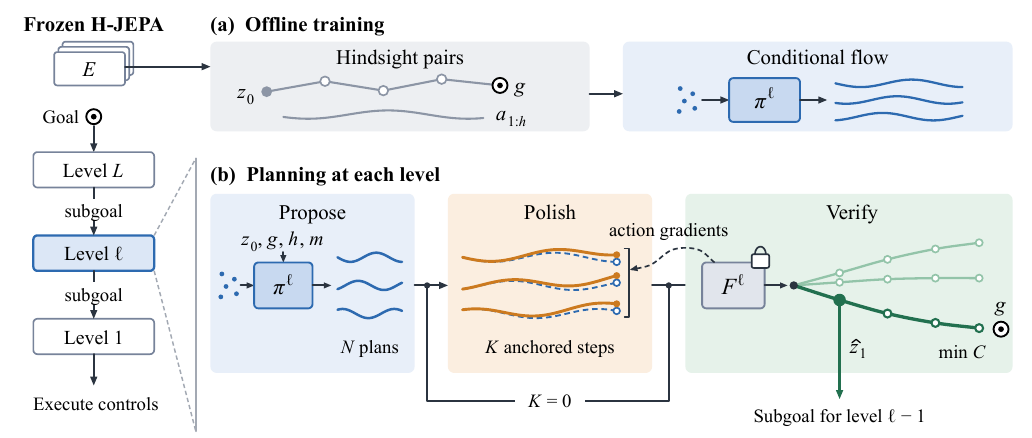}
\vspace{-0.5em}
\caption{\textbf{\pvp: proposals supply plans; the world model judges them.} \textbf{(a)}~Frozen encoders turn observed training trajectories into hindsight pairs; only the conditional flow heads are trained. \textbf{(b)}~At each level, the trained head proposes $N$ plans conditioned on state, target, horizon $h$, and target level $m$. $K$ anchored updates refine the entire pool through frozen-predictor action gradients. Final rollout selection uses the H-JEPA cost $C$; $K=0$ bypasses refinement. The chosen upper-level plan's first predicted state becomes the next subgoal, while the bottom level executes controls. Curves are schematic.}
\label{fig:method}
\vspace{-0.4em}
\end{figure*}

\section{CriticJEPA}
\label{sec:method}

\pvp generates goal-directed alternatives at every temporal level and uses frozen world-model predictions to choose among them. Short anchored updates refine every candidate before selection in the gradient-based variant. \Cref{fig:method} shows the computation, with its execution order in \cref{tab:alg}.

At level $\ell$, the frozen world model has latent state $z_t^{(\ell)}$, predictor $F^{(\ell)}$, and temporal stride $\Delta_\ell$. A plan $a_{1:h_\ell}^{(\ell)}$ contains control blocks at the bottom level and learned macro-actions above it. The predictor rolls out $\hat z_k^{(\ell)}=F^{(\ell)}(z_0^{(\ell)},a_{1:k}^{(\ell)})$. A target $g^{(m)}$ can arrive from level $m\geq\ell$. Frozen upper encoders project the rollout into that target space using the H-JEPA temporal windows, yielding $\tilde z_k^{(m)}$; no projection is needed when $m=\ell$. 

\paragraph{Propose: condition actions on state, target, and timescale.}
Several action sequences can connect a state to the same target. We learn a distribution of plans conditioned on the goal and timescale, using a flow proposer at each level~\cite{lipman2023flow,liu2023rectified}:
\begin{equation}
\begin{aligned}
a_{1:h_\ell}^{(\ell)}&\sim\pi_\theta^{(\ell)}(\,\cdot\mid c),\quad m\geq\ell,\\
c&=\big(z_0^{(\ell)},z_0^{(m)},g^{(m)}-z_0^{(m)},h_\ell,m\big).
\end{aligned}
\label{eq:cond}
\end{equation}
One sample produces an entire level-$\ell$ plan spanning $h_\ell\Delta_\ell$ primitive steps. The condition supplies the current state in the planning and target spaces, the displacement to the goal, and embeddings for the horizon and target level. When horizon decay skips an upper level, the H-JEPA rule targets the goal image in the skipped upper space for AntMaze/FourRoom and in the active level's space for Cube/Push-T. An AntMaze level-1 proposer can thus receive a level-3 goal while still producing a short control block.

We train on the world model's reward-free training trajectories using hindsight goals~\cite{andrychowicz2017her,ghosh2021gcsl}. A real action segment $x_1=a_{t:t+h-1}^{(\ell)}$ is paired with its starting latent and the latent it actually reaches after $h\Delta_\ell$ primitive steps. Short segments are padded and masked. For Gaussian $x_0$ and uniform $\tau\in[0,1]$, we interpolate $x_\tau=(1-\tau)x_0+\tau x_1$ and train a conditional velocity field:
\begin{equation}
\mathcal L_{\mathrm{flow}}^{(\ell)}
=\mathbb E\!\left[
\big\|v_\theta^{(\ell)}(x_\tau,\tau,c)-(x_1-x_0)\big\|_{\mathrm{valid}}^2
\right].
\label{eq:flow}
\end{equation}
The squared error uses the valid, unpadded action entries. Hindsight pairs connect each starting state to an observed outcome. Independent noise samples generate alternatives for the critic to compare.

All encoders, action encoders, and predictors remain frozen. At inference, four Euler steps generate $N_\ell$ candidates in parallel, clipped to the H-JEPA action bounds.

\paragraph{Verify: use the world model as a rollout critic.}
Once proposals are available, the frozen predictor chooses among them using its relative judgments. For the current state, \mbox{target}, and upper-level plan, we retain the \mbox{H-JEPA rollout cost}:
\begin{equation}
\begin{aligned}
C^{(\ell)}(a)
 &= \sum_{k\in\mathcal T_\ell}\!\big\|\tilde z_k^{(m)}-g^{(m)}\big\|_2^2+\beta_\ell R_{\mathrm{act}}^{(\ell)}(a),\\
i^\star &= \argmin_{i=1,\ldots,N_\ell} C^{(\ell)}(a^{(i)}).
\end{aligned}
\label{eq:cost}
\end{equation}
Here $\mathcal T_\ell$ contains the final scoring steps used by H-JEPA. The action-consistency term $R_{\mathrm{act}}^{(\ell)}$ sums squared differences between actions encoded into the upper action space and the selected upper plan over their matching window. Its weight is $\beta_1=0.2$ in Cube and Push-T when an upper action target is present, and zero otherwise. The highest level targets the encoded goal image.

For \textbf{\pvp-0}, one batched evaluation selects the lowest-cost proposal. Candidates run in parallel, while each rollout advances through the horizon. Higher levels pass the selected plan's first predicted state downward; the bottom level executes its control block. This forward selection forms a complete planner throughout the hierarchy, using the original dynamics and temporal abstraction.

\paragraph{Polish: anchor short updates to the proposals.}
The early gains in the iteration probe motivate short local refinement of the proposed plans. For \textbf{\pvp-$K$}, we retain all $N_\ell$ proposals and refine them in parallel for $K$ AdamW steps before the final critic selection. Let $a_{\mathrm{prop}}^{(i)}$ be proposal $i$ and $d_a$ the number of action entries. We optimize
\begin{equation}
\mathcal J^{(\ell)}(a^{(i)})=
C^{(\ell)}(a^{(i)})+
\frac{\lambda}{d_a}\|a^{(i)}-a_{\mathrm{prop}}^{(i)}\|_2^2,
\label{eq:polish}
\end{equation}
with $\lambda=0.1$ in the normalized action space. The anchor penalizes departures from the proposal while allowing local corrections. Final selection uses the rollout cost $C^{(\ell)}$ without the anchor penalty: the anchor shapes local refinement, while the critic compares the resulting plans. This gives test-time compute two distinct roles. Candidate count controls the breadth of comparison by supplying alternatives to the critic; polish steps control local adaptation around the original proposals.

\begin{table}[t]
\centering
\small
\caption{\textbf{One hierarchy, two modes.} $K=0$ uses forward verification alone; $K>0$ polishes the full candidate set before selection. Target-space projections, horizon decay, and execution windows follow H-JEPA.}
\label{tab:alg}
\vspace{-0.4em}
\setlength{\tabcolsep}{2pt}
\renewcommand{\arraystretch}{1.08}
\begin{tabular}{@{}>{\scriptsize}r>{\raggedright\arraybackslash}p{0.925\linewidth}@{}}
\toprule
\multicolumn{2}{@{}l}{\textbf{Algorithm 1:} \pvp at each replanning window}\\
\midrule
 & \textbf{Input:} frozen hierarchy, proposers $\pi_\theta^{(\ell)}$, counts $N_\ell$, horizons $h_\ell$, polish steps $K$\\
1 & Encode the current observation at all levels; initialize target $g\leftarrow E^{(L)}(o_g)$ and target level $m\leftarrow L$\\
2 & \textbf{for} $\ell=L,\ldots,1$ \textbf{do}\\
3 & \quad \textbf{if} $\ell>1$ and $h_\ell=1$: skip this level using the H-JEPA goal-space rule; continue\\
4 & \quad Form condition $c$ from \cref{eq:cond}; sample $N_\ell$ plans from $\pi_\theta^{(\ell)}(\cdot\mid c)$\\
5 & \quad If $K>0$, polish \emph{all} plans for $K$ steps on \cref{eq:polish}\\
6 & \quad Verify all final plans; select $i^\star=\argmin_i C^{(\ell)}(a^{(i)})$\\
7 & \quad If $\ell>1$, set $g\leftarrow F^{(\ell)}(z_0^{(\ell)},a_1^{(i^\star)})$ and $m\leftarrow\ell$\\
8 & Execute the selected bottom-level plan for the H-JEPA execution window and replan\\
\bottomrule
\end{tabular}
\vspace{-0.5em}
\end{table}

\paragraph{Where planning computation goes.}
Zero-gradient planning uses four proposer evaluations and one predictive rollout per candidate. Each polish step adds a rollout and its action-gradient computation. This applies to macro-actions above and control blocks below. Including proposer cost, the savings come from replacing repeated world-model optimization with a short generative pass and a final comparison.

\section{Experiments}
\label{sec:exp}

\subsection{Experimental Setup}
\label{sec:setup}

\paragraph{Benchmarks and frozen world models.}
We evaluate pretrained H-JEPA hierarchies~\cite{zhang2026hjepa} across four control domains, using three independently trained world models for each environment and hierarchy depth. \textbf{Visual AntMaze} uses a three-level model for quadruped maze navigation from $64{\times}64$ top-down pixels and proprioception, with a 350-step budget. \textbf{Push-T} uses two- and three-level models for contact-rich planar pushing of a T-block to a target pose, with a 90-step budget. \textbf{OGBench Cube} uses both hierarchy depths for 6-DoF robotic cube manipulation~\cite{park2025ogbench}, with a 30-step budget. \textbf{FourRoom} uses a three-level model for multi-room navigation through narrow doorways, with a 210-step budget. All 18 world models remain frozen; each model's flow proposers learn from its own encoded training trajectories. On Visual AntMaze, three planning restarts on each of the 50 official tasks give 450 episodes per method across the three models. Diagnostics and component ablations hold one AntMaze model fixed to isolate changes in planning. A further 50 tasks generated after method development form the \emph{Unseen} set, with 150 episodes per method. Each remaining setting uses 50 tasks per world model, also giving 150 episodes per method.

\paragraph{Methods and metrics.}
The official \textbf{H-JEPA} gradient planner runs up to 90 AdamW steps per level with its original early stopping. To separate generation from selection, we include \textbf{Proposal}, a single-proposal ablation that executes one deterministic flow sample per level at each replanning step without rollout verification. \textbf{\pvp-0} verifies $N_\ell$ candidates in one batched rollout; \textbf{\pvp-$K$} applies $K$ anchored refinement steps to every candidate before selection. On AntMaze the H-JEPA candidate counts are $100/100/400$: L1 samples a standard Gaussian, and upper levels sample Gaussians fitted to training macro-actions with H-JEPA action clipping. We report success as mean $\pm$ standard error across the three world models and median environment steps in successful episodes. Planner TFLOPs per episode count encoder, proposer, and predictor operations, including refinement backward passes.

\paragraph{Training and shared planning settings.}
The velocity field is a six-block residual MLP of width 512 with AdaLN conditioning~\cite{peebles2023dit}. Training uses 25 epochs, batch size 1024, and a cosine learning-rate schedule starting at $3\!\times\!10^{-4}$; the final epoch is used for every model.

All experiments retain the H-JEPA hierarchy, horizon schedule, and execution windows. Stochastic proposals use temperature 1.0; the deterministic ablation starts the flow at zero noise. We fix four Euler steps and $\lambda=0.1$. The cross-environment comparison uses $N_1/N_2/N_3=100/100/200$ for \pvp-0 and $25/25/100$ with $K=5$ for \pvp-5, using the first two counts for two-level models. We establish this configuration on one AntMaze model and transfer it unchanged to the other models, environments, and unseen tasks. The AntMaze budget sweep varies only the stated candidate counts and polish steps.

\subsection{Main Results on Visual AntMaze}
\label{sec:antmaze}

\begin{table}[t]
\centering
\normalsize
\caption{\textbf{Visual AntMaze: higher success with lower planning compute.} Three frozen world models, 450 Standard and 150 Unseen episodes per row. Success is in percent, mean $\pm$ SE across models; Steps is the median among successful episodes. Candidate counts $N_1/N_2/N_3$ are $8/8/16$ for the small variant, $100/100/200$ for \pvp-0, and $25/25/100$ for \pvp-5. TFLOPs is mean model compute per Standard episode.}
\label{tab:antmaze}
\vspace{-0.4em}
\renewcommand{\arraystretch}{1.10}
\setlength{\tabcolsep}{3pt}
\begin{tabular}{@{}lrrrr@{}}
\toprule
Method & Standard & Unseen & TFLOPs & Steps \\
\midrule
H-JEPA & 67.6\ds{2.5} & 74.7\ds{2.4} & 131.0 & 226 \\
Single proposal & 44.7\ds{4.1} & 45.3\ds{3.5} & 0.25 & 226 \\
\midrule
\pvp-0 (small) & 68.7\ds{1.7} & 78.7\ds{3.5} & \textbf{0.36} & 267 \\
\pvp-0 & 78.7\ds{4.8} & 87.3\ds{2.7} & 1.62 & 234 \\
\pvp-5 & \textbf{89.1}\ds{2.1} & \textbf{92.0}\ds{1.2} & 5.11 & \textbf{199} \\
\bottomrule
\end{tabular}
\vspace{-0.5em}
\end{table}

\begin{figure}[t]
\centering
\includegraphics[width=\linewidth]{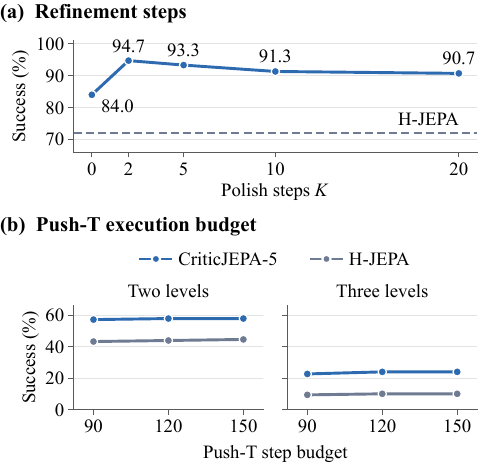}
\vspace{-0.5em}
\caption{\textbf{Success peaks with short refinement; gains persist across execution budgets.} \textbf{(a)}~The AntMaze sweep holds the model and $25/25/100$ candidate counts fixed, averaging three restarts on 50 tasks. \textbf{(b)}~Push-T success across three independently trained models, with 150 episodes per curve. The same runs continue beyond the official 90-step budget at the minimum horizon, retaining the official horizon schedule through step 90.}
\label{fig:polish}
\vspace{-0.6em}
\end{figure}

Frozen rollout selection turns generated alternatives into more successful control. In \cref{tab:antmaze}, selecting from a proposal pool raises Standard success from 44.7\% for a single proposal to 78.7\% for \pvp-0. The latter reaches 87.3\% on Unseen tasks at 1.62 TFLOPs per episode. These gains of 11.1 and 12.6 percentage points over H-JEPA use the same frozen dynamics. The small pool reaches 68.7\% at 0.36 TFLOPs, exceeding H-JEPA's 67.6\% at 131.0 TFLOPs with about $360\times$ less planning compute.

Short refinement provides a stronger operating point with fewer candidates. \pvp-5 reaches 89.1\% success at 5.11 TFLOPs, a gain of 21.5 percentage points over H-JEPA with $26\times$ less planning compute. Success improves for every independently trained world model. The same proposal heads and frozen predictors reach 92.0\% on Unseen tasks, compared with H-JEPA's 74.7\%, carrying the gain to new start--goal pairs. Successful episodes also finish sooner, with a median of 199 steps versus 226. This configuration combines a smaller pool with five anchored updates; \cref{fig:polish}a follows refinement length at a fixed pool size.

The same trained proposer supports two ways of spending planning compute: generating more alternatives and adapting those already proposed. Two-step refinement reaches 88.2\% at 2.51 TFLOPs, while five steps with a larger $50/50/200$ pool reach 90.9\% at 9.97 TFLOPs. Across the AntMaze budget sweep, planning compute ranges from 0.36 to 18.63 TFLOPs per episode without retraining.

\subsection{Across Domains and Hierarchy Depths}
\label{sec:cross}

\begin{table}[t]
\centering
\small
\caption{\textbf{Shared settings improve success and cut compute across domains and depths.} Success in percent, reported as mean $\pm$ standard error over three independently trained models per row. \emph{Proposal} executes a single proposal per level without verification. \emph{Compute reduction} divides H-JEPA planner compute by \pvp-5 compute.}
\label{tab:cross}
\vspace{-0.4em}
\renewcommand{\arraystretch}{1.16}
\setlength{\tabcolsep}{1.35pt}
\begin{tabular}{@{}llccccr@{}}
\toprule
Env. & Levels & H-JEPA & Proposal & \multicolumn{2}{c}{\pvp} & Compute \\
 & & & & $K=0$ & $K=5$ & reduction \\
\midrule
Cube & $L{=}2$ & 49.3\ds{8.4} & \textbf{100}\ds{0.0} & \textbf{100}\ds{0.0} & 92.0\ds{1.2} & 44$\times$ \\
Cube & $L{=}3$ & 48.7\ds{4.1} & \textbf{100}\ds{0.0} & \textbf{100}\ds{0.0} & \textbf{100}\ds{0.0} & 50$\times$ \\
Push-T & $L{=}2$ & 43.3\ds{1.8} & 34.7\ds{4.8} & 52.0\ds{3.1} & \textbf{57.3}\ds{1.3} & 79$\times$ \\
Push-T & $L{=}3$ & 9.3\ds{2.9} & 18.0\ds{2.0} & 18.7\ds{1.8} & \textbf{22.7}\ds{1.3} & 13$\times$ \\
FourRoom & $L{=}3$ & 96.0\ds{1.2} & \textbf{100}\ds{0.0} & \textbf{100}\ds{0.0} & \textbf{100}\ds{0.0} & 14$\times$ \\
AntMaze & $L{=}3$ & 67.6\ds{2.5} & 44.7\ds{4.1} & 78.7\ds{4.8} & \textbf{89.1}\ds{2.1} & 26$\times$ \\
\bottomrule
\end{tabular}
\vspace{-0.5em}
\end{table}

\begin{table*}[t]
\centering
\small
\caption{\textbf{Hierarchical proposals combine control gains and compute savings.} All comparisons share the diagnostic AntMaze model from \cref{sec:probing} and three restarts on 50 tasks; \cref{tab:antmaze} averages over three models. Success is percent; TFLOPs is mean per episode. In (a), $N_1/N_2/N_3$ is $1/1/1$ for single proposals, $100/100/200$ for \pvp-0, and $25/25/100$ for \pvp-5. In (b), Upper denotes L2/L3; H-JEPA denotes gradient search. The $K=0$ columns form a matched $2\times2$ comparison. At $K=10$, L1-only and all-level proposals use 100 and 50 L1 candidates.}
\label{tab:ablation}
\vspace{-0.4em}
\begin{minipage}[t]{0.485\textwidth}
\centering
\textit{(a) Planning components}\par
\setlength{\tabcolsep}{2pt}
\begin{tabular}{@{}lccrr@{}}
\toprule
\multirow{2}{*}{Configuration} & \multirow{2}{*}{Verify} & \multirow{2}{*}{Polish $K$} & \multicolumn{2}{c}{Performance} \\
 & & & Success & TFLOPs \\
\midrule
H-JEPA & -- & -- & 72.0 & 131.6 \\
Proposal, stochastic & -- & -- & 33.3 & 0.25 \\
Proposal, deterministic & -- & -- & 52.0 & 0.25 \\
\pvp-0 & $\checkmark$ & 0 & 86.7 & 1.62 \\
\pvp-5 & $\checkmark$ & 5 & \textbf{93.3} & 5.11 \\
\bottomrule
\end{tabular}
\end{minipage}\hfill
\begin{minipage}[t]{0.485\textwidth}
\centering
\textit{(b) Proposal placement}\par
\setlength{\tabcolsep}{2pt}
\begin{tabular}{@{}llrrrr@{}}
\toprule
\multirow{2}{*}{Upper} & \multirow{2}{*}{L1} & \multicolumn{2}{c}{$K=0$} & \multicolumn{2}{c}{$K=10$} \\
 & & Success & TFLOPs & Success & TFLOPs \\
\midrule
H-JEPA & H-JEPA & 72.0 & 131.6 & 72.0 & 131.6 \\
Proposals & H-JEPA & 63.3 & 15.86 & 76.0 & 31.59 \\
H-JEPA & Proposals & 84.7 & 112.96 & 88.7 & 115.71 \\
Proposals & Proposals & \textbf{86.7} & \textbf{1.62} & \textbf{90.7} & \textbf{18.63} \\
\bottomrule
\end{tabular}
\end{minipage}
\vspace{-0.5em}
\end{table*}

The same candidate budgets and refinement settings carry the AntMaze gains to the other domains and hierarchy depths. Across all six settings in \cref{tab:cross}, \pvp-5 improves on H-JEPA while cutting planning compute by $13\times$ to $79\times$. The tasks reveal distinct roles for generation, selection, and local refinement.

On Cube and FourRoom, the generator supplies successful behavior directly. A single deterministic proposal solves \textbf{100\%} of episodes on every world model at both Cube depths and on three-level FourRoom. \pvp-0 preserves full success without gradient refinement; H-JEPA reaches 49.3\% and 48.7\% on two- and three-level Cube. With smaller candidate pools, \pvp-5 reaches 92.0\% on two-level Cube, a 42.7-point gain over H-JEPA at $44\times$ lower compute, and 100\% on three-level Cube and FourRoom.

Push-T, like AntMaze, benefits from having alternatives to compare. On two-level Push-T, generating and verifying a candidate pool raises success from 34.7\% for a single proposal to 52.0\%. The five-step configuration reaches 57.3\% at $79\times$ lower compute, compared with H-JEPA's 43.3\%. On three-level Push-T, where short episodes restrict valid pretraining clips~\cite{zhang2026hjepa}, \pvp-5 more than doubles H-JEPA success, from 9.3\% to 22.7\%. Extending the same runs from 90 to 150 environment steps preserves the advantage in \cref{fig:polish}b. With 67\% more execution steps, H-JEPA success changes by at most 1.3 points, while \pvp-5 maintains higher success across all budgets.

\subsection{Understanding the Planning Gains}
\label{sec:analysis}

\paragraph{From a learned action to a choice of plans.}
On the diagnostic AntMaze model, single stochastic and deterministic proposals solve 33.3\% and 52.0\% of tasks, respectively, as shown in \cref{tab:ablation}a. Generating multiple alternatives and selecting by frozen rollout scores raises success to 86.7\%. The learned distribution supplies the candidate behaviors, and comparison adapts the choice to the current state and goal. With a smaller pool and five anchored updates, \pvp-5 reaches 93.3\% at 5.11 TFLOPs per episode.

\paragraph{The hierarchy links control quality to search cost.}
Applying the design at both timescales brings together higher success and lower compute. In the $K=0$ comparison of \cref{tab:ablation}b, replacing L1 search raises success from 72.0\% to 84.7\%. The upper-level optimizers still consume most of the computation, leaving the total at 112.96 TFLOPs per episode. With L1 proposals in place, replacing upper-level search further raises success to 86.7\% and cuts compute to 1.62 TFLOPs, a $70\times$ reduction. The lower level improves the actions executed; the upper levels make the resulting controller inexpensive to run.

Upper-level proposals are most effective with proposal-based control. With H-JEPA at L1, they reach 63.3\% at $K=0$. Their $K=10$ configuration reaches 76.0\% at 31.59 TFLOPs, surpassing H-JEPA's 72.0\% with about one quarter of the compute. At $K=10$, the full hierarchy reaches 90.7\% at 18.63 TFLOPs, compared with 88.7\% at 115.71 TFLOPs for L1-only proposals. Pairing proposals at every level delivers the highest success and lowest compute at both refinement budgets.

\paragraph{Short corrections improve complete-task success.}
Short refinement captures the largest gains in complete-task success. With the world model and candidate counts fixed, two anchored steps raise success from 84.0\% to \textbf{94.7\%}, as shown in \cref{fig:polish}a. Ten and twenty steps retain 91.3\% and 90.7\% success; the peak is reached with two updates. The three-model AntMaze sweep shows the same preference for short refinement: with $50/50/200$ candidates, five steps achieve 90.9\% and ten achieve 89.8\%. Gradients are most effective here as local corrections to generated plans, while the frozen critic makes the final choice. This complete-task profile connects the planner's compute allocation to the early execution gains in \cref{sec:exploit}.

The same frozen hierarchy also succeeds more consistently across restarts. \pvp-5 averages 93.3\% success per run, exceeding the 92.0\% coverage obtained by combining three H-JEPA restarts in \cref{sec:gap}. Three runs of \pvp-10 solve \textbf{all 50 tasks} in their union, and the fraction of tasks whose outcomes change across those runs drops from 48\% to 22\%.

\section{Discussion and Conclusion}
\label{sec:conclusion}

Frozen predictions support better plan selection after continued gradient search stops improving execution. Exact-state AntMaze execution reveals this distinction in the same frozen predictor: its rankings distinguish goal-conditioned proposals and are uninformative for H-JEPA search outputs. The candidate source determines both which plans the model compares and what its selected action achieves.

\pvp turns this insight into a complete hierarchical planner. Flow heads learn goal-directed alternatives from reward-free trajectories, frozen rollouts choose among them, and short anchored updates capture the early gains from gradients before selection. This division operates on both macro-actions and primitive controls. Forward selection preserves full success on Cube and FourRoom, where proposals already supply task-solving behavior. On AntMaze and Push-T, comparing alternatives and applying short corrections yield further gains from the same frozen predictors. Shared settings carry these gains across independently trained world models, hierarchy depths, and unseen tasks. Every world-model weight stays fixed.

Planning compute now serves two complementary roles: generating alternatives and locally refining them before frozen rollout selection. Candidate generation shapes what can be achieved, while the model's comparisons determine which action is executed. This division recovers more successful behavior from unchanged world models.

\par

\bibliographystyle{ieeenat_fullname}
\bibliography{main}

\begin{thebibliography}{33}
\providecommand{\natexlab}[1]{#1}
\providecommand{\url}[1]{\texttt{#1}}
\expandafter\ifx\csname urlstyle\endcsname\relax
  \providecommand{\doi}[1]{doi: #1}\else
  \providecommand{\doi}{doi: \begingroup \urlstyle{rm}\Url}\fi

\bibitem[Ajay et~al.(2023)Ajay, Du, Gupta, Tenenbaum, Jaakkola, and
  Agrawal]{ajay2023decisiondiffuser}
Anurag Ajay, Yilun Du, Abhi Gupta, Joshua~B. Tenenbaum, Tommi~S. Jaakkola, and
  Pulkit Agrawal.
\newblock Is conditional generative modeling all you need for decision-making?
\newblock In \emph{International Conference on Learning Representations}, 2023.

\bibitem[Andrychowicz et~al.(2017)Andrychowicz, Wolski, Ray, Schneider, Fong,
  Welinder, McGrew, Tobin, Abbeel, and Zaremba]{andrychowicz2017her}
Marcin Andrychowicz, Filip Wolski, Alex Ray, Jonas Schneider, Rachel Fong,
  Peter Welinder, Bob McGrew, Josh Tobin, Pieter Abbeel, and Wojciech Zaremba.
\newblock Hindsight experience replay.
\newblock In \emph{Advances in Neural Information Processing Systems}, 2017.

\bibitem[Assran et~al.(2023)Assran, Duval, Misra, Bojanowski, Vincent, Rabbat,
  LeCun, and Ballas]{assran2023ijepa}
Mahmoud Assran, Quentin Duval, Ishan Misra, Piotr Bojanowski, Pascal Vincent,
  Michael Rabbat, Yann LeCun, and Nicolas Ballas.
\newblock Self-supervised learning from images with a joint-embedding
  predictive architecture.
\newblock In \emph{IEEE/CVF Conference on Computer Vision and Pattern
  Recognition}, 2023.

\bibitem[Assran et~al.(2025)Assran, Bardes, Fan, Garrido, Howes,
  et~al.]{assran2025vjepa2}
Mido Assran, Adrien Bardes, David Fan, Quentin Garrido, Russell Howes, et~al.
\newblock {V-JEPA} 2: Self-supervised video models enable understanding,
  prediction and planning.
\newblock \emph{arXiv preprint arXiv:2506.09985}, 2025.

\bibitem[Balestriero and LeCun(2025)]{balestriero2025lejepa}
Randall Balestriero and Yann LeCun.
\newblock {LeJEPA}: Provable and scalable self-supervised learning without the
  heuristics.
\newblock \emph{arXiv preprint arXiv:2511.08544}, 2025.

\bibitem[Bansal et~al.(2026)Bansal, Naphade, Aggarwal, and
  M]{bansal2026leplanner}
Saksham Bansal, Om Naphade, Chayan Aggarwal, and Vrishin M.
\newblock {LePlanner}: An iterative amortized controller for world models.
\newblock \emph{arXiv preprint arXiv:2609.13845}, 2026.

\bibitem[Bar et~al.(2025)Bar, Zhou, Tran, Darrell, and LeCun]{bar2025nwm}
Amir Bar, Gaoyue Zhou, Danny Tran, Trevor Darrell, and Yann LeCun.
\newblock Navigation world models.
\newblock In \emph{IEEE/CVF Conference on Computer Vision and Pattern
  Recognition}, 2025.

\bibitem[Bardes et~al.(2024)Bardes, Garrido, Ponce, Chen, Rabbat, LeCun,
  Assran, and Ballas]{bardes2024vjepa}
Adrien Bardes, Quentin Garrido, Jean Ponce, Xinlei Chen, Michael Rabbat, Yann
  LeCun, Mahmoud Assran, and Nicolas Ballas.
\newblock Revisiting feature prediction for learning visual representations
  from video.
\newblock \emph{arXiv preprint arXiv:2404.08471}, 2024.

\bibitem[Caselli et~al.(2026)Caselli, Massafra, Punzo, Sardo, Pantelidis, and
  Bhethanabhotla]{caselli2026mindthegap}
Niccol{\`o} Caselli, Francesco Massafra, Samuele Punzo, Salvatore~Lo Sardo,
  Ippokratis Pantelidis, and Sathya~Kamesh Bhethanabhotla.
\newblock Mind the gap: Promises and pitfalls of hierarchical planning in
  {LeWorldModel}.
\newblock \emph{arXiv preprint arXiv:2607.12547}, 2026.

\bibitem[Cheng et~al.(2026)Cheng, Zhang, and Wang]{cheng2026sage}
Letian Cheng, Qi Zhang, and Yisen Wang.
\newblock {SAGE}: Subgoal-conditioned action generation for latent world model
  planning.
\newblock \emph{arXiv preprint arXiv:2607.17973}, 2026.

\bibitem[Chi et~al.(2023)Chi, Xu, Feng, Cousineau, Du, Burchfiel, Tedrake, and
  Song]{chi2023diffusionpolicy}
Cheng Chi, Zhenjia Xu, Siyuan Feng, Eric Cousineau, Yilun Du, Benjamin
  Burchfiel, Russ Tedrake, and Shuran Song.
\newblock Diffusion policy: Visuomotor policy learning via action diffusion.
\newblock In \emph{Robotics: Science and Systems}, 2023.

\bibitem[Chua et~al.(2018)Chua, Calandra, McAllister, and Levine]{chua2018pets}
Kurtland Chua, Roberto Calandra, Rowan McAllister, and Sergey Levine.
\newblock Deep reinforcement learning in a handful of trials using
  probabilistic dynamics models.
\newblock In \emph{Advances in Neural Information Processing Systems}, 2018.

\bibitem[Ghosh et~al.(2021)Ghosh, Gupta, Reddy, Fu, Devin, Eysenbach, and
  Levine]{ghosh2021gcsl}
Dibya Ghosh, Abhishek Gupta, Ashwin Reddy, Justin Fu, Coline Devin, Benjamin
  Eysenbach, and Sergey Levine.
\newblock Learning to reach goals via iterated supervised learning.
\newblock In \emph{International Conference on Learning Representations}, 2021.

\bibitem[Hafner et~al.(2019)Hafner, Lillicrap, Fischer, Villegas, Ha, Lee, and
  Davidson]{hafner2019planet}
Danijar Hafner, Timothy Lillicrap, Ian Fischer, Ruben Villegas, David Ha,
  Honglak Lee, and James Davidson.
\newblock Learning latent dynamics for planning from pixels.
\newblock In \emph{International Conference on Machine Learning}, 2019.

\bibitem[Hansen et~al.(2024)Hansen, Su, and Wang]{hansen2024tdmpc2}
Nicklas Hansen, Hao Su, and Xiaolong Wang.
\newblock {TD-MPC2}: Scalable, robust world models for continuous control.
\newblock In \emph{International Conference on Learning Representations}, 2024.

\bibitem[Janner et~al.(2019)Janner, Fu, Zhang, and Levine]{janner2019mbpo}
Michael Janner, Justin Fu, Marvin Zhang, and Sergey Levine.
\newblock When to trust your model: Model-based policy optimization.
\newblock In \emph{Advances in Neural Information Processing Systems}, 2019.

\bibitem[Janner et~al.(2022)Janner, Du, Tenenbaum, and
  Levine]{janner2022diffuser}
Michael Janner, Yilun Du, Joshua~B. Tenenbaum, and Sergey Levine.
\newblock Planning with diffusion for flexible behavior synthesis.
\newblock In \emph{International Conference on Machine Learning}, 2022.

\bibitem[LeCun(2022)]{lecun2022path}
Yann LeCun.
\newblock A path towards autonomous machine intelligence.
\newblock OpenReview preprint, 2022.

\bibitem[Lipman et~al.(2023)Lipman, Chen, Ben-Hamu, Nickel, and
  Le]{lipman2023flow}
Yaron Lipman, Ricky T.~Q. Chen, Heli Ben-Hamu, Maximilian Nickel, and Matt Le.
\newblock Flow matching for generative modeling.
\newblock In \emph{International Conference on Learning Representations}, 2023.

\bibitem[Liu et~al.(2023)Liu, Gong, and Liu]{liu2023rectified}
Xingchao Liu, Chengyue Gong, and Qiang Liu.
\newblock Flow straight and fast: Learning to generate and transfer data with
  rectified flow.
\newblock In \emph{International Conference on Learning Representations}, 2023.

\bibitem[Loshchilov and Hutter(2019)]{loshchilov2019adamw}
Ilya Loshchilov and Frank Hutter.
\newblock Decoupled weight decay regularization.
\newblock In \emph{International Conference on Learning Representations}, 2019.

\bibitem[Maes et~al.(2026)Maes, Lidec, Scieur, LeCun, and
  Balestriero]{maes2026lewm}
Lucas Maes, Quentin~Le Lidec, Damien Scieur, Yann LeCun, and Randall
  Balestriero.
\newblock {LeWorldModel}: Stable end-to-end joint-embedding predictive
  architecture from pixels.
\newblock \emph{arXiv preprint arXiv:2603.19312}, 2026.

\bibitem[Park et~al.(2025)Park, Frans, Eysenbach, and Levine]{park2025ogbench}
Seohong Park, Kevin Frans, Benjamin Eysenbach, and Sergey Levine.
\newblock {OGBench}: Benchmarking offline goal-conditioned {RL}.
\newblock In \emph{International Conference on Learning Representations}, 2025.

\bibitem[Peebles and Xie(2023)]{peebles2023dit}
William Peebles and Saining Xie.
\newblock Scalable diffusion models with transformers.
\newblock In \emph{IEEE/CVF International Conference on Computer Vision}, 2023.

\bibitem[Sobal et~al.(2025)Sobal, Zhang, Cho, Balestriero, Rudner, and
  LeCun]{sobal2025pldm}
Vlad Sobal, Wancong Zhang, Kyunghyun Cho, Randall Balestriero, Tim G.~J.
  Rudner, and Yann LeCun.
\newblock Learning from reward-free offline data: A case for planning with
  latent dynamics models.
\newblock \emph{arXiv preprint arXiv:2502.14819}, 2025.

\bibitem[Sun et~al.(2026)Sun, Zhao, and Zhang]{sun2026intact}
Junhan Sun, Hao Zhao, and Guofeng Zhang.
\newblock {INTACT}: Isomorphic intent-to-action learning for search-free world
  models.
\newblock \emph{arXiv preprint arXiv:2607.26056}, 2026.

\bibitem[Szegedy et~al.(2014)Szegedy, Zaremba, Sutskever, Bruna, Erhan,
  Goodfellow, and Fergus]{szegedy2014intriguing}
Christian Szegedy, Wojciech Zaremba, Ilya Sutskever, Joan Bruna, Dumitru Erhan,
  Ian Goodfellow, and Rob Fergus.
\newblock Intriguing properties of neural networks.
\newblock In \emph{International Conference on Learning Representations}, 2014.

\bibitem[Terver et~al.(2026)Terver, Yang, Ponce, Bardes, and
  LeCun]{terver2026what}
Basile Terver, Tsung-Yen Yang, Jean Ponce, Adrien Bardes, and Yann LeCun.
\newblock What drives success in physical planning with joint-embedding
  predictive world models?
\newblock \emph{Transactions on Machine Learning Research}, 2026.

\bibitem[Wang and Ba(2020)]{wang2020poplin}
Tingwu Wang and Jimmy Ba.
\newblock Exploring model-based planning with policy networks.
\newblock In \emph{International Conference on Learning Representations}, 2020.

\bibitem[Wang et~al.(2026)Wang, Xia, Zhou, Hu, Shi, Du, and Ye]{wang2026prism}
Yuhai Wang, Jiawei Xia, Rongxuan Zhou, Xiao Hu, Yongliang Shi, Jing Du, and
  Yang Ye.
\newblock {PRISM}: {PR}ior-guided imagination sampling in world models.
\newblock \emph{arXiv preprint arXiv:2606.07974}, 2026.

\bibitem[Zhang et~al.(2026{\natexlab{a}})Zhang, Terver, Rabbat, LeCun, and
  Balestriero]{zhang2026hjepa}
Wancong Zhang, Basile Terver, Michael Rabbat, Yann LeCun, and Randall
  Balestriero.
\newblock {H-JEPA}: End-to-end learning of hierarchical world models for visual
  planning.
\newblock \emph{arXiv preprint arXiv:2610.06805}, 2026{\natexlab{a}}.

\bibitem[Zhang et~al.(2026{\natexlab{b}})Zhang, Terver, Zholus, Chitnis,
  Sutaria, Assran, Balestriero, Bar, Bardes, LeCun, et~al.]{zhang2026hwm}
Wancong Zhang, Basile Terver, Artem Zholus, Soham Chitnis, Harsh Sutaria, Mido
  Assran, Randall Balestriero, Amir Bar, Adrien Bardes, Yann LeCun, et~al.
\newblock Hierarchical planning with latent world models.
\newblock \emph{arXiv preprint arXiv:2604.03208}, 2026{\natexlab{b}}.

\bibitem[Zhou et~al.(2025)Zhou, Pan, LeCun, and Pinto]{zhou2025dinowm}
Gaoyue Zhou, Hengkai Pan, Yann LeCun, and Lerrel Pinto.
\newblock {DINO-WM}: World models on pre-trained visual features enable
  zero-shot planning.
\newblock In \emph{International Conference on Machine Learning}, 2025.

\end{thebibliography}

\flushcolsend
\end{document}